\documentclass{article}

\usepackage{PRIMEarxiv}

\usepackage[utf8]{inputenc} 
\usepackage[T1]{fontenc}    
\usepackage{hyperref}       
\usepackage{url}            
\usepackage{booktabs}       
\usepackage{amsfonts}       
\usepackage{nicefrac}       
\usepackage{microtype}      
\usepackage{lipsum}
\usepackage{algorithm}
\usepackage{algpseudocode}
\usepackage{fancyhdr}       
\usepackage{graphicx} 
\usepackage{tikz}
\graphicspath{{media/}}     
\usepackage[english]{babel}
\usepackage{amsthm}
\newtheorem{theorem}{Theorem}
\newtheorem{corollary}{Corollary}[theorem]
\newtheorem{lemma}[theorem]{Lemma}
\theoremstyle{definition}
\newtheorem{definition}{Definition}[section]
\title{Online Fair Revenue Maximizing Cake Division with Non-Contiguous Pieces in Adversarial Bandits
\thanks{\textit{\underline{Citation}}: 
\textbf{Mohammad Ghodsi, Amirmahdi Mirfakhar}} 
}

\author{
  Mohammad Ghodsi\\
  Professor \\
  Sharif University of Technology \\
  Tehran\\
  \texttt{ghodsi@sharif.edu} \\
   \And
  Amirmahdi Mirfakhar \\
  Master of Science Student \\
  Sharif University of Technology \\
  Tehran\\
  \texttt{amir.mirfakhar@sharif.edu} \\
}

\begin{document}
\maketitle

\begin{abstract}
The classic cake-cutting problem provides a model for addressing the fair and efficient allocation of a divisible, heterogeneous resource among agents with distinct preferences. Focusing on a standard formulation of cake cutting, in which each agent must receive a contiguous piece of the cake in an offline setting, this work instead focuses on online allocating non-contiguous pieces of cake among agents and establishes algorithmic results for fairness measures. In this regard, we made use of classic adversarial multi-armed bandits to achieve sub-linear Fairness and Revenue Regret at the same time. Adversarial bandits are powerful tools to model the adversarial reinforcement learning environments, that provide strong upper-bounds for regret of learning with just observing one action's reward in each step by applying smart trade-off between exploration and exploitation. This work studies the power of the famous EXP\_3 algorithm that is based on exponential wight{-}importance updating probability distribution through time horizon.
\end{abstract}

\keywords{cake division \and fairness \and multi-armed bandits \and adversarial bandits \and sub-linear regret}

\section{Introduction}
Cake-cutting is a fundamental problem in the literature of fairness. It is about allocating a divisible resource among agents with different preferences over different parts of it. Indeed, the classic work of Steinhaus, Banach, and Knaster \cite{steihaus1948problem} which lays the mathematical foundations
of fair division—addresses cake cutting. Throughout the years, this problem has not only inspired many people to develop new mathematical and algorithmic approaches but also found its way into real-world problems like sharing borders and natural resources.

In formal form, the cake is represented by segment $[0,1]$, and the preferences of the agents are shown by functions over the cake's sub-intervals. The cake owner takes a knife and cuts it into pieces, then shares the pieces among the agents. There are generally two different cake division approach, in one of them the cake owner divides the cake into $n$ pieces and divides it among $n$ agents which is called cake-cutting with contiguous pieces. Whereas, in the second approach, cake-cutting with contiguous pieces, the owner divides the cake into more than or equal to $n$ pieces and then allocates them among agents; Obviously, it is possible that an agent receives more than one piece which might be separate intervals of the cake \cite{brams1996fair}.

Achieving fairness and efficiency are two pivotal goals in this resource-allocation context. The current work contributes to one of these objectives, with a focus on computational aspects of cake-cutting. The fairness and efficiency objectives addressed in this work are detailed next.

A basis for the fairness notion is envy-freeness: a division is called envy-free if and only if, under it, every agent prefers its own piece over that of any other agent. It has been proved that under mild assumptions on agents’ valuations, envy-free cake divisions with connected pieces always exist \cite{edward1999rental}. But there is another study that proves to find the envy-free solution cannot be computed in bounded time if the valuations are specified by an adaptive adversary \cite{stromquist2008envy}.

All previously mentioned settings of the cake-cutting problem were defined in offline settings, in which there exists an oracle that tells the cake owner each agent's valuation. In other words, the cake owner knows each agent's valuation overall sub-intervals of $[0,1]$. However, in the real world, the agent's valuations change rapidly over time and the cake owner does not know the agents' real valuations over different cuts of the cake. This natural assumption is considered in this work. Here, there is a cake which is allocated through time. The agents also would have different valuations over cake at different times. In the current work, the cake owner runs an online algorithm in adversarial bandit setting to learn near-optimal envy-free allocation through time by estimating the agent's valuations at each step.

Bandit algorithms (Bubeck \& Cesa-Bianchi \cite{bubeck2012regret}; Lattimore \&
Szepesvari \cite{lattimore2020bandit}) provide an attractive model of learning for online platforms, and they are now widely used to optimize retail, media streaming, and news feed. Each round of bandit learning corresponds to an interaction with a user, where the algorithm selects an arm (e.g. product, song, article, agent to receive a piece of a cake) observes the user’s response (e.g. purchase, stream, read, agent's valuation over given piece),
and then updates its policy. Over time, the bandit algorithm
thus learns to maximize the user responses, which are often
well aligned with the objective of the online platform (e.g.
profit maximization, engagement maximization, fairness). While maximizing users' profit, bandit algorithms might neglect fair and deserved exposure to every arm. There are not many studies which have addressed this concern in their works. Recent work from Wang, Bai, Sun \& Joachim \cite{wang2021fairness} addressed fairness in expected number of exposure to each arm in stochastic bandits which altered famous UCB algorithm to guarantee sub-linear regret from optimal exposure to each arm with respect to its importance. Our work defines a new setting in adversarial bandits for online cake-cutting with non-contiguous pieces with defining two regrets to compete with at the same time: revenue and fairness. Revenue Regret relates to the cake owner and Fairness Fegret concerns about difference between agents' accumulated valuations through time.

In bandit settings, we generally try to devise efficient algorithms to achieve sub-linear regret. Lattimore \& Szepesvari \cite{lattimore2020bandit} have proposed algorithms in both adversarial and stochastic bandits, which receive sub-linear upper bound $ C \sqrt {n k \log (k)}$ for Revenue Regret where $n$ is the horizon for a multi-armed bandit with $k$ arms. Achieving sub-linear Revenue and Fairness Regret upper bound at the same time would be Obviously harder. Although we will show that achieving sub-linear upper bounds for both regrets simultaneously could be impossible, we will make use of famous EXP\_3 algorithms to achieve revenue and Fairness Regrets in an environment where achieving both is possible. First, we propose two algorithms to reach sub-linear regret for revenue and fairness separately. Then we merge two algorithms to achieve both sub-linear upper bounds at the same time. Finally, we will generalize a framework, in which EXP\_3 can help us to reach our ultimate goal and achieve sub-linear upper bounds for revenue and Fairness Regret.

\subsection{Related Works}
Algorithmic fairness has been extensively studied in binary classification (Hardt et al., 2016 \cite{hardt2016equality}; Chouldechova, 2017 \cite{chouldechova2017fair}; Kleinberg et al., 2017; \cite{kleinberg2016inherent}). These works propose statistical criteria to test algorithmic fairness that sometimes exploits definitions of fairness from political philosophy and sociology. Several prior works like Blum et al.,
2018 \cite{blum2018preserving} and Blum \& Lykouris, 2019 \cite{blum2019advancing} study how to achieve these fairness criteria in online learning. These algorithms achieve fairness to the incoming users like what we do here.

Joseph et al. \cite{joseph2016fair} study fairness
in bandits that ensure a better arm is always selected with no less probability than a worse arm. Different from our definition of fairness, their optimal policy is still the one that deterministically selects the arm with the largest expected reward while giving zero exposure to all the other arms.

Our definition of fairness has connections to the fair division problem, which was defined by Procaccia \cite{procaccia2013cake} where the goal is to allocate a resource to different agents in a fair way. In our problem, we aim to allocate the users’ attention among the items in a fair way. In our setting someone cuts the cake and the online bandit algorithm decides who to allocate the cut piece. The agent who achieves the cut piece of cake will pay something to the cake owner. Here, we will devise an algorithm to find a Fairness Regret suffering a sub-linear bound and use another algorithm to find a sub-linear bound for the Revenue Regret. Finally, we study environments where the latter one can achieve both sub-linear bounds at the same time.

As far as we know, this work is the first one that analyzes the famous envy-free cake-cutting problem as a multi-armed bandit problem with revenue concerns.

\section{Notation and Preliminaries}
We consider the problem of dividing a cake (which metaphorically represents a divisible, heterogenous good) among $n$ agents. In this setup, the cake is modeled as the segment $[0, 1]$ and the (possibly) distinct cardinal preferences of the agents are expressed as valuation functions, $\left\{v_{a}\right\}_{a \in[n]}$, over the intervals contained in $[0, 1]$ (i.e., over the pieces of the cake).

Every agent should pay $\left\{p_{a}\right\}_{a \in[n]}$ to the cake owner, which is considered to be his/her valuation over the received piece of the cake multiplied by a constant factor. In other word, if agent $a$ receives the piece of cake $[l,r]$, then should pay $p_{a}([l,r]) = \alpha.v_{a}([l,r])$. Here, for simplicity, we assume $\alpha = 1$.

There is also another important assumption here, that is similar to that of the conventional cake-cutting problem. At every round $t$, all agents' valuations over the remaining cake is less than one. To clarify, we have  $p_{a}([r_{t-1},1])_{a \in[n]} \leq 1$. 
\subsection{Adversarial Bandit Setting}
At first, we assume that the agent's preferences over different parts of the cake changes over time. In other words, the time variable would be entered into the preference function of each agent. The value of the agent $a$ at time step $t$ for the $[l,r]$ piece of cake is shown by $v_{a}([l,r])$ and the payment for the receiver would be $p_{a}([l,r])$.

To transform the offline cake-cutting problem into the online one, we assume that at each step the adversary cuts a piece of cake and the cake owner should allocate that cut piece to one of the agents. The agent then pays equal to its valuation to the cake owner. Cutting by the adversary is equivalent to the setting that the adversary chooses agents' valuations over the constant piece of cake. Thus, we can assume that the cake owner cuts a piece of cake to allocate at each round and the adversary specifies each agent's valuation and relatively the payment. Achieving the payment, the cake owner (the learner) earns the payment and finds out the valuation of This game repeats for $T$ rounds till the cake owner allocates the whole cake.

In the translation into the Adversarial Bandit's setting, the cake owner is the learner and each of the $n$ agents is one of the arms. Throughout the horizon of $T$ rounds, at round $t$, the learner cuts the piece $[l_t,r_t]$ of the cake and pulls the arm $a_t$ which means allocating the $[l_t,r_t]$ to the agent $a_t$. Then the learner observes the payment $p_{a_t}([l_t,r_t])$ which is set by an adversary.

In adversarial settings, the adversary completely knows the learner's algorithm, thus the learner should use a randomized algorithm to be able to achieve sub-linear regret. In each round, the learner chooses a distribution over the
actions $\mathcal{P}_{t}$. Then the action $a_t$ is sampled from $\mathcal{P}_{t}$, and the learner receives reward $p_{a_t}([l_t,r_t])$. A policy in this setting is a function $\pi:([n] \times [0,1])^{*}$ mapping history sequences to distributions over actions (according to the assumption the reward of pulling each arm at every round is less than or equal to 1).

\subsubsection{Revenue Regret}
Here we define a Revenue Regret, which is exactly the utility regret of conventional adversarial bandits. This term measures the difference between the performance of the cumulative revenue of the cake owner through $T$ rounds of exploration and exploitation with the accumulated revenue of the policy of allocating the cut piece of the cake to one agent at every step, which achieves the maximum accumulated revenue.

$$
RevR_{T}(\pi, p)=\max _{i \in[n]} \sum_{t=1}^{T} p_{i}([l_t,r_t])-\mathbb{E}\left[\sum_{t=1}^{T} p_{a_t}([l_t,r_t])\right] \label{rregret:Revenue}
$$

\subsubsection{Fairness Regret}
This regret is our innovation that separates this work from the previous ones. Here, the concern is accumulated rewards that each arm has provided. When it comes to cake-cutting language we want to compare how much cake have agents received. In this regard, we define the fairness measure of an allocation policy, as the difference between the highest accumulated valuation among all agents and the lowest of the others.

$$
F(\pi,T) = \max _{a \in[n]} \sum_{t=1}^{T} v_{a}([l_t,r_t]) \cdot x_{a, t}-\min _{b \in[n]} \sum_{t=1}^{T} v_{b}([l_t,r_t]) \cdot x_{b, t}
$$

Where $\pi$ is the policy, and $x_{a, t}$ is an indicator variable which is equal to $1$ if the agent $a$ achieves $[b_t,c_t]$ and $0$ otherwise. If $F(\pi,T)=0$ the cake has been allocated envy-freely.

Fairness Regret is defined based on $F(\pi,T)$. This regret measures the linear distance of the policy $\pi$'s fairness and the optimal offline algorithm's fairness which can allocate $[l_t,r_t]$s without chronological order to achieve optimal fairness. We name this offline algorithm $OPT$. So the Fairness Regret will be:

$$FairR_{T}(\pi, p)=\mathbb{E}[F(\pi,T)] - F(OPT,T)$$\label{fregret:Fairness}

\section{Independent Algorithms}
\subsection{Revenue Maximizing Algorithm}
As we mentioned before, the algorithm that the learner should make use of to be able to offset the adversary should be randomized. Here we make use of a weight-importance randomized algorithm with an unbiased estimator for estimating other agents' valuation after observing that of just one agent. Then we use the exponential weighting method to produce a probability distribution over the agents at each step to achieve sub-linear Revenue Regret in adversarial bandit setting.

\subsubsection{Importance-Weighted Estimators}
A key ingredient of all adversarial bandit algorithms is a mechanism for estimating the reward of unplayed arms. Recall that $\mathcal{P}_{t}$ is the conditional distribution of the action played in round $t$, and so for $i \in[n], P_{t i}$ is the conditional probability
$$
\mathcal{P}_{t i}=\mathbb{P}\left(a_{t}=i \mid a_{1}, p_{a_1}([l_1,r_1]), \ldots, a_{t-1}, p_{a_{t-1}}([l_{t-1},r_{t-1}])\right)
$$
In what follows, we assume that for all $t$ and $i, P_{t i}>0$ almost surely. As we shall see later, this will be true for all policies considered in this chapter. The importance-weighted estimator of $p_{i}([l_t,r_t])$ is
$$
\hat P_{i}([l_t,r_t])=\frac{\mathbb{I}\left\{a_{t}=i\right\} p_{a_t}([l_t,r_t])}{\mathcal{P}_{t i}}
$$

It would be easy to check that $\mathbb{E}\left[\hat P_{i}([l_t,r_t])\right] = p_{i}([l_t,r_t])$.

To continue, we change our estimator to be able to make use of conventional adversarial bandit regret bound to achieve sub-linear Revenue Regret.

According to the assumption that every agents' valuation over the $[l_t,r_t]$ is less than $1$, We change our focus from feeding the estimator with the reward ($p_{a_t}([l_t,r_t])$) to feeding it with the $L_{a_t}([l_t,r_t])$ which is defined as

$$Y_{a_t}([l_t,r_t])= 1 - p_{a_t}([l_t,r_t])$$

Then we subtract every reward dependent variable from the previous estimator to devise new one that is build upon Loss instead of Reward. The new estimator would be

$$
\hat Y_{i}([l_t,r_t])=\frac{\mathbb{I}\left\{a_{t}=i\right\} Y_{a_t}([l_t,r_t])}{\mathcal{P}_{t i}}
$$

We did this because by the nature of the agent's valuation which is less than $1$, we can decrease the variance of our estimator by this slight change.

To convert it into reward dependent estimator (just for readability) we re-write it as

$$
\hat P_{i}([l_t,r_t])=1-\frac{\mathbb{I}\left\{A_{t}=i\right\}}{{\mathcal{P}_{t i}}}\left(1-p_{a_t}([l_t,r_t])\right)
$$

\subsubsection{Exponential Weighting Distribution}
Let $\hat{S}_{t i}=\sum_{s=1}^{t} \hat P_{i}([l_s,r_s])$ be the total estimated reward by the end of round $t$ It seems natural to play actions with larger estimated reward with higher probability. While there are many ways to map $\hat{S}_{t i}$ into probabilities, a simple and popular choice is called exponential weighting, which for tuning parameter $\eta > 0$ sets

$$
\mathcal{P}_{t i}=\frac{\exp \left(\eta \hat{S}_{t-1, i}\right)}{\sum_{j=1}^{n} \exp \left(\eta \hat{S}_{t-1, j}\right)}
$$
The parameter $\eta$ is called the learning rate and we will tune it by choosing $\eta$ to depend only on the number of agents $n$.

\subsubsection{Revenue Maximizing EXP\_3}
The algorithm which is called EXP\_3 will be

\begin{algorithm}
\caption{Revenue Maximizing EXP\_3}\label{alg:cap}
\begin{algorithmic}
\State Set $\hat{S}_{0 i} = 0$ for all $i$
\For {$t = 1,2,...,T$}
\State Calculate the sampling distribution $\mathcal{P}_{t}$:
\State $$
\mathcal{P}_{t i}=\frac{\exp \left(\eta \hat{S}_{t-1, i}\right)}{\sum_{j=1}^{n} \exp \left(\eta \hat{S}_{t-1, j}\right)}
$$
\State Sample $a_{t} \sim \mathcal{P}_{t}$ and observe reward $p_{a_t}([l_t,r_t])$
\State Calculate $\hat{S}_{t i}$:
$$
\hat{S}_{t i}=\hat{S}_{t-1, i}+1-\frac{\mathbb{I}\left\{A_{t}=i\right\}}{{\mathcal{P}_{t i}}}\left(1-p_{a_t}([l_t,r_t])\right)
$$
\EndFor
\end{algorithmic}
\end{algorithm}

\subsubsection{Regret Analysis}
To analyze the Revenue Regret and prove a sub-linear bound for it we use the theorem below which is the result of Lattimore \& Szepesvari \cite{lattimore2020bandit} for similar adversarial bandits.

\begin{theorem}\label{theorem:theorem1}
Let $P \in[0,1]^{T \times n}$ and $\pi$ be the policy of Exp\_3 \ref{alg:cap} with learning rate $\eta=\sqrt{\log (n) /(T n)}$. Then,
$$
RevR_{T}(\pi, p) \leq 2 \sqrt{T n \log (n)}
$$
\end{theorem}

The result of \ref{theorem:theorem1} proves that the EXP\_3 algorithm could guarantee a sub-linear upper bound for the Revenue Regret.

\subsection{Fairness Minimizing Algorithm}
Here, first we reduce our online fair cake-cutting problem to a famous problem which is known to be $NP\textunderscore Hard$. Then we specify the algorithm that achieves the $OPT$ in Fairness Regret \ref{fregret:Fairness} for having a practical regret for future experiments. However the algorithm we propose will achieve constant Fairness Regret regardless of what algorithm we utilize to find $OPT$ for comparing our online algorithm with.

\subsubsection{Online Job Scheduling Problem}
In \textbf{Job Scheduling} or \textbf{Minimum Makespan Scheduling Problem} defined as follows: Suppose we have $k$ jobs each of which take time $t_{i}$ to process, and $m$ identical machines on which we schedule the jobs. Jobs cannot be split between machines. For a given scheduling, let $Q_{j}$ be the set of jobs assigned to machine $j .$ Let $JST_{j}=\sum_{i \in Q_{j}} t_{i}$ be the load of machine $j .$ The minimum makespan scheduling problem asks for an assignment of jobs to machines that minimizes the makespan, where the makespan is defined as the maximum load over all machines (i.e. $\max _{j} JST_{j}$ ).

This problem has both offline and online setting. In the offline mode the set of $k$ jobs are given but in the online setting jobs arrive chronologically at time steps and each of them should be assigned to one of the machines instantly.

It has been proved that the offline Job Scheduling Problem is \emph{NP\_Hard}. It also isn't arduous to admit that the online setting is harder.

\subsubsection{Reduction}

There is a harder version of Job Scheduling Problem where each job $i$ has $t_{i,j}$ running time on machine $j$. In other words, each job would have different running times on different machines. We name this version \textbf{Real Job Scheduling Problem}.\label{real:RJSP}

Here we reduce our online fair cake-cutting problem to the Real Job Scheduling Problem. To do this, first we can consider $[l_t,r_t]$ as a job $i$ which arrives at time $t$. The agent $a$ serves as machine $j$, thus if we put the $t_{i,j}$ equal to $p_{a}([l_t,r_t])$ our setting would be similar to the Real Job Scheduling.

To prove that our problem is at least as hard as the Real Job Scheduling Problem, first we should prove the following theorem:

\begin{theorem}\label{theorem:theorem2}
Reducing the online fair cake-cutting problem to the Real Scheduling Problem, minimizing the $\max _{j} JST_{j}$ is not harder than minimizing the Fairness Regret under no assumptions on valuations.
\begin{proof}
If we relax the assumption of having different running times on different machines we would have $t_{i,j} = t_{i,k}$ for all $j$ and $k$. This is equivalent to the online cake-cutting problem where every agent has equal valuation over $[l_t,r_t]$. Under this circumstance, it is easy to check that minimizing the $\max _{j} JST_{j}$, which is known to be $NP\textunderscore Hard$, would be equal to minimizing the Fairness Regret. If we again assume that $t_{i,j}$ are different, both of these would be considerably harder.
\end{proof}
\end{theorem}

\subsubsection{Allocate to Min Algorithm}
Although there is a LP-based $2-approx$ algorithm for offline Real Job Scheduling Problem, there is no simple and straight-forward approximation algorithm for Online fair cake-cutting problem. But as we know that $OPT \geq 0$ and $p_{a_t}([l_t,r_t]) \leq 1$ for all $a$ and $t$s, we propose a straight forward algorithm to achieve constant regret according to the fantasy $OPT=0$ algorithm.
Here we define $AT_{a,t}$ as $\sum_{i=1}^{t} v_{a_i}([l_i,r_i]).x_{a,i}$, which is total valuation that agent $a$ has achieved from round $1$ to $t$.
The algorithm is called allocate to min:
\begin{algorithm}
\caption{Allocate to Min}\label{alg2:caption}
\begin{algorithmic}
\State Set $AT_{a,0} = 0$ for all $a$
\For {$t = 1,2,...,T$}
\State $a_{min}= argmin_{a\in[n]} AT_{a,t}$
\State allocate $[l_t,r_t]$ to $a_{min}$
\State $AT_{a_{min},t} = AT_{a_{min},{t-1}} + v_{a_{min}}([l_t,r_t])$
\EndFor

\end{algorithmic}
\end{algorithm}

\subsubsection{Regret Analysis}
To do the Regret Analysis we need to prove this theorem:

\begin{theorem}\label{theorem:theorem2}
Using the Allocate to Min we would have
$$FairR_{T}(Allocate\ to\ Min, p) \leq 1$$

\begin{proof}
We know, from our main assumption, that the valuation of every agents on $[l_t,r_t]$ would be less than or equal to $1$ at each round $t$. Therefore after first round the $FairR_{1}(Allocate\ to\ Min, p)$ would be at most $1$. Since the first round, allocating the $[l_t,r_t]$ to $a_{min}$ would reduce the regret till the agent with the maximum accumulated valuation changes and the other appears in Fairness Regret. If we assume that the maximum change occurs at time $i$ and agent with maximum accumulated valuation at time $i-1$ as $a_{max_{i-1}}$ and minimum as $a_{min_{i-1}}$, it would be easy to prove that the $a_{min_{i-1}} = a_{max_{i}}$ and $$AT_{a_{min_{i-1}},i-1} + v_{a_{min_{i-1}}}([l_{i-1},r_{i-1}]) - v_{a_{min_{i}}} \leq 1  $$
To continue we can do like everything has been reset to the beginning and $FairR_{i}(Allocate\ to\ Min, p)$ which is now less than or equal to $1$, decreases again to the next maximum change. Therefore, using simple induction would prove the theorem. 
\end{proof}
\end{theorem}

The result is amazing, because we have devised a simple online algorithm which achieves a constant Fairness Regret.

\section{Fair Revenue Maximizing Algorithm}
At this part, we know that we have two separate algorithms to guarantee sub-linear bounds for both Revenue and Fairness Regret. Our next step would be trying to find an algorithm to satisfy both demands over sub-linearity of Revenue and Fairness Regret at the same time.
\subsection{Impossibility}
Here we prove that achieving sub-linear regret for both of the regrets would be impossible in adversarial bandits. In other words, there exist cutting series $[l_t,r_t]$, in which the adversary can force us to miss the sub-linearity over one of our regrets while we strive to achieve the other. The following theorem illustrates the impossibility of gaining both sub-linear regrets at the same time

\begin{theorem}\label{theorem:theorem2}
According to the Revenue and Fairness Regret we defined in \ref{fregret:Fairness} and \ref{rregret:Revenue}, it is impossible to have an algorithm which can achieve sub-linear bound for Revenue and Fairness Regret in adversarial setting at the same time.

\begin{proof}
to prove the theorem, we construct an example to show the impossibility:

\begin{tikzpicture}
\draw[step=0.5cm,gray,very thin] (0,2.5) rectangle (16.5,8.5);
\draw[step=0.5cm,gray,very thin] (0,2.5) grid (16.5,8.5);
\filldraw[draw=none, fill=white] (0,7.5) rectangle node{1} (2,8);
\filldraw[fill=blue!40!white, draw=black] (2,7.5) rectangle node{$v_{1}(c_1)$} (4,8);
\filldraw[fill=blue!40!white, draw=black] (4,7.5) rectangle node{$v_{1}(c_2)$} (6,8);
\filldraw[fill=blue!40!white, draw=black] (6,7.5) rectangle node{$v_{1}(c_3)$} (8,8);
\filldraw (8.75,7.75) circle (0.125cm);
\filldraw (9.25,7.75) circle (0.125cm);
\filldraw (9.75,7.75) circle (0.125cm);
\filldraw[fill=blue!40!white, draw=black] (10.5,7.5) rectangle node{$v_{1}(c_{T-1})$} (12.5,8);
\filldraw[fill=blue!40!white, draw=black] (12.5,7.5) rectangle node{$v_{1}(c_T)$} (14.5,8);

\filldraw[draw=none, fill=white] (0,6.5) rectangle node{2} (2,7);
\filldraw[fill=blue!40!white, draw=black] (2,6.5) rectangle node{$v_{2}(c_1)$}  (2.90,7);
\filldraw[fill=blue!40!white, draw=black] (2.90,6.5) rectangle node{$v_{2}(c_2)$} (3.8,7);
\filldraw[fill=blue!40!white, draw=black] (3.8,6.5) rectangle node{$v_{2}(c_3)$} (4.7,7);
\filldraw (4.95,6.75) circle (0.0625cm);
\filldraw (5.2,6.75) circle (0.0625cm);
\filldraw (5.45,6.75) circle (0.0625cm);
\filldraw[fill=blue!40!white, draw=black] (5.7,6.5) rectangle node{$v_{2}(c_T)$} (6.6,7);

\filldraw[draw=none, fill=white] (0,5.5) rectangle node{3} (2,6);
\filldraw[fill=blue!40!white, draw=black] (2,5.5) rectangle node{$v_{3}(c_1)$}  (2.90,6);
\filldraw[fill=blue!40!white, draw=black] (2.90,5.5) rectangle node{$v_{3}(c_2)$} (3.8,6);
\filldraw[fill=blue!40!white, draw=black] (3.8,5.5) rectangle node{$v_{3}(c_3)$} (4.7,6);
\filldraw (4.95,5.75) circle (0.0625cm);
\filldraw (5.2,5.75) circle (0.0625cm);
\filldraw (5.45,5.75) circle (0.0625cm);
\filldraw[fill=blue!40!white, draw=black] (5.7,5.5) rectangle node{$v_{3}(c_T)$} (6.6,6);

\filldraw (5.2,4.75) circle (0.0625cm);
\filldraw (5.2,4.5) circle (0.0625cm);
\filldraw (5.2,4.25) circle (0.0625cm);

\filldraw[draw=none, fill=white] (0,3) rectangle node{n} (2,3.5);
\filldraw[fill=blue!40!white, draw=black] (2,3) rectangle node{$v_{n}(c_1)$}  (2.90,3.5);
\filldraw[fill=blue!40!white, draw=black] (2.90,3) rectangle node{$v_{n}(c_2)$} (3.8,3.5);
\filldraw[fill=blue!40!white, draw=black] (3.8,3) rectangle node{$v_{n}(c_3)$} (4.7,3.5);
\filldraw (4.95,3.25) circle (0.0625cm);
\filldraw (5.2,3.25) circle (0.0625cm);
\filldraw (5.45,3.25) circle (0.0625cm);
\filldraw[fill=blue!40!white, draw=black] (5.7,3) rectangle node{$v_{n}(c_T)$} (6.6,3.5);
\end{tikzpicture}

In this example we assume that the adversary chooses $v_{i}(c_t)$ for each agent where $c_t = [l_t,r_t]$. The adversary here, puts $v_1(c_t)\approx1$ and other valuations $v_a(c_t)=\epsilon\approx0$ for all $t$ and $a\in\{2, 3, \dots, n\}$. In this example, achieving sub-linear Revenue Regret like $f(T)$ needs to allocate $[l_t,r_t]$ for at least $T-f(T)$ times to the agent $1$. This means that other agents will receive at most $T.\epsilon \approx 0$. Therefore, the Fairness Regret would be at least $T-f(T)-T.\epsilon$ which is linear. Achieving sub-linear Fairness Regret is equal to allocate $[l_t,r_t]$ to the agent $1$ for sub-linear times which contradicts achieving sub-linear bound for the Revenue Regret.
\end{proof}
\end{theorem}

\subsection{Fine Environment}
To be able to reach the goal of finding an algorithm that provides both Revenue and Fairness Regret, we should first limit the adversary to commit such cuttings on the cake which makes achieving sub-linear Revenue and Fairness Regret Possible. Thus, in the remaining parts, we assume that the adversary would cut pieces of cake in such a way that sub-linear regrets are achievable. In other words, we try to compete with weaker adversaries that we call \textbf{Fine Environment}.
\subsection{Fair Revenue Maximizing EXP\_3 (FRM EXP\_3)}
It has been proved by Lattimore \& Szepesvari \cite{lattimore2020bandit} that our only existing approach to overcome the adversary to achieve sub-linear bound for the Revenue Regret is to use EXP\_3-like algorithms with such estimators that we have used. Thus, having in our minds that EXP\_3 guarantees sub-linear Revenue Regret, we focus on finding  Fine Environments when the EXP\_3 with slight change can guarantee the sub-linear Fairness Regret too.

\subsubsection{$cf(t)$-Domination}
Here we need to define a term to be able to describe the Fine Environments where FRM EXP\_3 can guarantee achieving sub-linear Fairness Regret. 
\begin{definition}[$cf(t)$-Domination]
Under time cut $[t_1,t_2]$ and a constant $c$, we say that the agent $a$, $cf(t)$-dominates the agent $b$, if $(\sum_{t=t_1}^{t_2} v_{a}([l_t,r_t]) - \sum_{t=t_2}^{t_2} v_{b}([l_t,r_t])) \in c.\mathcal{O}(f(t))$, where $f(t) \in o(t_2-t_1)$.
\end{definition}

We can also use the Domination term on the sets of agents similarly.
\begin{definition}
We say a set $A$ of agents, $cf(t)$-dominates a set $B$ of agents where $A\cap B = \emptyset$ and each agent $a \in A$, $cf(t)$-dominates every agent $b\in B$.
\end{definition}

\subsubsection{FRM EXP\_3}
At this section we slightly change EXP\_3 algorithm to be more compatible with Fairness Regret bounding.
\begin{algorithm}
\caption{FRM EXP\_3}\label{alg3:caption}
\begin{algorithmic}
\State Set $\theta = \frac {1}{\sqrt{T}}$
\State Set $AT_{a,0} = 0$ for all $a$
\For {$t = 1,2,...,T$}
\State Follow EXP\_3 with probability $(1-\theta)$ or Allocate to Min with probability $(\theta)$

\EndFor

\end{algorithmic}
\end{algorithm}

FRM EXP\_3 is still guaranteed to achieve sub-linear Revenue Regret because the new algorithms commit $\sqrt{T}$, expectedly, to Allocate to Min algorithm. Therefore, it is expected to lose $\sqrt{T}$ of possible revenue which means the Revenue Regret will remain sub-linear.
\subsubsection{Fine Environment's Attributes for FRM EXP\_3 }
Here we describe Fine Environments for FRM EXP\_3 and prove that it achieves sub-linear Fairness Regret under such circumstances. At first, we clarify how the EXP\_3-like algorithms achieve sub-linear Revenue Regret, compare the best agent in revenue making through history. EXP\_3-like algorithms on our cake-cutting problem strive to incline to the subset of agents where they $cf(t)$-dominate other agents in revenue making. Thus, we prove the following theorem for the first attribute of Fine Environments for FRM EXP\_3 :

\begin{theorem}\label{theorem:theorem5}
If we have such an environment where no sets of agents $cf(T)$-dominates the other agents, the FRM EXP\_3 algorithm would achieve sub-linear bound for Fairness Regret.

\begin{proof}
Under such circumstance, having a time cut $[t_1,t_2]$ where $t_1 \geq 0, t_2 \leq T$ and $t_2-t_1 \in \mathcal{O}(T)$ with a set of agents $A$ that $cf(t_2-t_1)$-dominates other agents contradicts the assumption. Therefore, we can conclude that the adversary has cut cake in such a way that the agent's valuations is guaranteed to be quasi-uniform distributed. Accordingly, we can make use of the result in \cite{lattimore2020bandit} and conclude that the Exponential Weighting Distribution remains roughly the same for all agents. Which means that every on would achieve $\frac{1}{n} \mathbb{E}\left[\max _{i \in[n]} \sum_{t=1}^{T} v_{i}([l_t,r_t])\right] \pm g(T)$, where $g(T)$ is a sub-linear function in terms of $T$. So, here, the FRM EXP\_3 not only does achieve sub-linear bound for Revenue Regret, but also does the same on Fairness Regret.

\end{proof}
\end{theorem}

We simply proved that if valuations are distributed quasi-uniformly through time, the FRM EXP\_3. But the story is completely different in environments where there is a time cut $[t_1,t_2]$, where a set of agents $cf(T)$-dominates the others. The reason was mentioned before, the nature of EXP\_3-like algorithms looks for finding agents (arms) where their possible accumulated revenue are higher than the others in terms of $T$. Thus, in the remainder of this part, we will specify the characteristics of Fine Environments with dominant sets of agents where our FRM EXP\_3 algorithm would work as a fair-revenue maximizing algorithm.

Having agents who $cf(T)$-dominate other agents at specific time cuts is a natural assumption that has a reflection in the real-life where someone like the coming-to-allocate $[l_t,r_t]$ far more than the others at a specific time interval. Here we define Fine Environments with domination sets. Through the remaining proofs of this time, we just analyze a single agent's domination over the others, but by using the following corollary we can generalize our results to sets of agents' domination too.

\begin{corollary}\label{corollary:cor1}
We can translate the set $A$ of $k$ agents' $cf(T)$-domination at the time cut $[t_1,t_2]$ over the other agents to individual $\frac{c}{k}f(T)$-domination at this time interval because if an EXP\_3-like algorithms switches from a trend to the set $A$, all of the agents would receive similar accumulated valuations according to \cite{lattimore2020bandit}.  
\end{corollary}

Considering \ref{corollary:cor1}, $w.l.g$, we focus on individual domination at different time cuts. We start with proving the following lemma.

\begin{lemma}\label{lemma:lem1}
If FRM EXP\_3 algorithm begins to allocate all pieces from $[l_{t_1},r_{t_1}]$ to $[l_{t_2},r_{t_2}]$ to agent $a$ who $cf(t_2-t_1)$-dominates others at time cut $[t_1,t_2]$. Immediate sub-linear Fairness Regret would be achieved between agent $a$ and $b$, when there is time $t_3$ at which the agent $b$, $2cf(t_2-t_1)$-dominates other agents at time cut $[t_2,t_3]$.
\begin{proof}
If $t_1$ would be the beginning time of the first dominating interval(before that everything is quasi-uniform), it means that FRM EXP\_3 finds and commits to agent $a$ as the first trend. Due to the definition of domination and EXP\_3-like algorithms nature in bounding the Revenue Regret, the agent $a$ has received $(\sum_{t=t_1}^{t_2} v_{a}([l_t,r_t]))-g(t_2-t_1)$, where $g(t)$ is a sub-linear function in terms of $t$. According to the lemma, if there is a time $t_3$ where the agent $b$, $2cf(t_2-t_1)$-dominates other agents, the FRM EXP\_3 algorithms converges to commit on agent $b$ to achieve the sub-linear Revenue Regret bound. It would be not that hard to show that the algorithm will catch $(\sum_{t=t_2}^{t_3} v_{b}([l_t,r_t]))-g(t_2-t_1)$. To prove the allegation, we recall the Revenue Regret in \ref{rregret:Revenue}. The sub-linear Revenue Regret means that the EXP\_3 algorithm should gain cumulative revenue with sub-linear difference from $\max _{i \in[n]} \sum_{t=1}^{t_3} p_{i}([l_t,r_t])$ which is equal to the same difference from  $\max _{i \in[n]} \sum_{t=1}^{t_3} v_{i}([l_t,r_t])$. Here the $\max _{i \in[n]} \sum_{t=1}^{t_3} p_{i}([l_t,r_t])$ is equal to the $2cf(t_2-t_1) \pm g(t_3)$. Half of this value was achieved during $a$'s trend; so the other half should be achieved by a commit on agent $b$. Therefore, we conclude that $a$ has achieved $cf(t_2-t_1) \pm g^\prime(t_2)$ and $b$ be has done $cf(t_2-t_1) \pm g^\prime(t_3)$. This means what $a$ and $b$ has achieved till $t_3$ have sub-linear Fairness Regret comparing to each other.
\end{proof}
\end{lemma}

The \ref{lemma:lem1} is illustrative and insightful to understand how the EXP\_3-like algorithms behave. Then we use this lemma to find the Fine Environment's attributes while a set of dominating agents exist. Then we introduce a pattern for all Fine Environments for FRM EXP\_3 algorithm.

\begin{definition}[Fair Monotone Domination Sequence]
For the super time cut $[t_l,t_r]$, we define a Monotone Domination Sequence as a chain of trends on each of the agents that, $w.l.g$, starts from a trend on agent $1$ at $t_l$ and ends after a final trend on agent $n$ at $t_r$. We call it a Fair Monotone Domination Sequence, where agent with $i$th trend on the chain, $i.cf(t)$-dominates other agents on it.
\end{definition}

We clarify the definition with the following example:

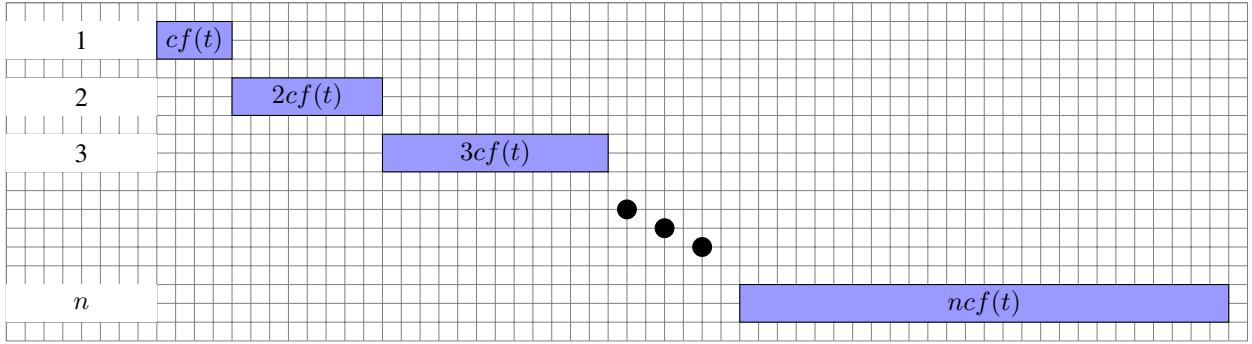
\begin{tikzpicture}
\draw[step=0.25cm,gray,very thin] (0,4) rectangle (16.5,8.5);
\draw[step=0.25cm,gray,very thin] (0,4) grid (16.5,8.5);

\filldraw[draw=none, fill=white] (0,7.75) rectangle node{1} (2,8.25);
\filldraw[fill=blue!40!white, draw=black] (2,7.75) rectangle node{$cf(t)$} (3,8.25);

\filldraw[draw=none, fill=white] (0,7) rectangle node{2} (2,7.5);
\filldraw[fill=blue!40!white, draw=black] (3,7) rectangle node{$2cf(t)$} (5,7.5);

\filldraw[draw=none, fill=white] (0,6.25) rectangle node{3} (2,6.75);
\filldraw[fill=blue!40!white, draw=black] (5,6.25) rectangle node{$3cf(t)$} (8,6.75);

\filldraw (8.25,5.75) circle (0.125cm);
\filldraw (8.75,5.5) circle (0.125cm);
\filldraw (9.25,5.25) circle (0.125cm);

\filldraw[draw=none, fill=white] (0,4.25) rectangle node{$n$} (2,4.75);
\filldraw[fill=blue!40!white, draw=black] (9.75,4.25) rectangle node{$ncf(t)$} (16.25,4.75);

\end{tikzpicture}
This example shows what Fair Monotone Domination Sequence is. The order of the agents who involve the trends doesn't have any effects on our general results. In this example, for simplicity, we assume that the trends follow the order $1, 2, \dots, n$.

At this part, we use the insight from the lemma \ref{lemma:lem1} to prove that at the end of all Fair Monotone Domination Sequence the Fairness Regret is sub-linear.

\begin{theorem}\label{theorem:theorem5}
In the environments where the domination intervals only follow the chain of Fair Monotone Domination Sequences pattern, the FRM EXP\_3 algorithm would achieve sub-linear Fairness Regret.

\begin{proof}
We prove the theorem for the first Fair Monotone Domination Sequence and the generalization to the chain of them would be done by induction. If a Fair Monotone Domination Sequence occurs for the first time at super time cut $[t_l,t_r]$, and $w.l.g$ the order of trends in that sequence be $1, 2, \dots, n$, the FRM EXP\_3 algorithm switches to trend of agent$1$ which $cf(t-t_1)$-dominates other agents. The FRM EXP\_3 was proved to achieve revenue $cf(t-t_1) - g(t-t_1)$ from the first agent to be compatible with its sub-linear Revenue Regret that was proved in \cite{lattimore2020bandit}. This is also equal to what agent $1$ gains as a cumulative value in it's trend. When trend changes to agent $2$ who $2cf(t-t_1)$-dominates others in his trend, the algorithm should receive another $cf(t-t_1) - g(t-t_1)$ from agent $2$ because the best agent through hindsight is changing from agent $1$ to agent $2$ and the algorithm should gain $2cf(t-t_1)$ to have sub-linear Revenue Regret. We know that $cf(t-t_1)$ of it was achieved at previous trend; therefore, the algorithms achieves another $cf(t-t_1)$ from the current trend. This process continues till the end of the sequence and algorithm guarantees to generate at least $ncf(t_2-t_1) - \sum_{i=1}^{n} g(t_2-c_i) $ revenue from all agents, where $c_i$ is the time that the $i$th trend starts. Through achieving this revenue, the algorithms makes $cf(t-t_1)- g(t-t_1)$ valuation for every agent at it's trend and it means agent's difference over their cumulative valuations would be sub-linear at time $t_2$. So under such circumstance, the sub-linear Fairness Regret is achieved.

\end{proof}
\end{theorem}

In sum, Fine Environments for the FRM EXP\_3 should have one or mixture of these two attributes:

\begin{enumerate}
    \item valuations should be quasi-uniformly distributed through time.
    \item dominating trends should follow the Fair Monotone Domination Sequences.
\end{enumerate}

Both of these attributes have a real reflection in the real life. Environments with quasi-uniform distributed valuations stand for the markets of well-established companies whose valuations never change considerably through time and they are at severe compete every day. And the second attribute stands for markets with old and new customers. Agents who start to bid for the resource should bid aggressively dominating to achieve what the old agents achieved before.

\section{Conclusion}
The current work studies online cake-cutting with adversarial bandits for the first time and proposes two separate algorithms for revenue and Fairness Regret. In particular, we develop two simple, efficient, and practical algorithms to find sub-linear regret bounds for two different objectives which make definite sense in real life. Then we try to find a unique and innovative algorithm to overcome two regret rivals at the same time. First, we prove the impossibility of achieving both sub-linear regrets simultaneously by constructing an illustrative example. Finally, we devise our last algorithm which is the mixture of our two separate algorithms for revenue-maximizing and fairness achieving. We complement our algorithm by introducing the tight and general attributes which the adversary should have to let our FRM EXP\_3 algorithm gain sub-linear Fairness Regret.
\section{Acknowledgements}
This work was inspired by our previous work with Professor Kasra Alishahi (Assistant Professor at Sharif Department of Mathematics) and Professor Mohammad-Hadi Foroughmand Arabi (Assistant Professor at Sharif Department of Computer Science).However, all content represents the opinion of the authors.

\bibliographystyle{unsrt}  
\bibliography{references}

\end{document}